\documentclass{article}
\usepackage{spconf,amsmath,graphicx,hyperref}

\usepackage{graphicx}
\usepackage{subcaption}
\usepackage{comment}
\usepackage{booktabs}
\usepackage{multirow} 

\usepackage{longtable}
\usepackage{xcolor}
\usepackage{algorithm}
\usepackage{algpseudocode}
\usepackage{setspace}
\usepackage{caption}
\usepackage{float} 
\usepackage[normalem]{ulem}

\newrobustcmd{\B}{\bfseries}
\newrobustcmd{\U}[1]{\uline{#1}}

\title{Quality-Aware Framework for Video-Derived Respiratory Signals}

\name{\begin{tabular}{c}Nhi Nguyen$^{\star}$, Constantino \'Alvarez Casado$^{\star}$$^{\dagger}$, \\ Le Nguyen$^{\star}$, Manuel Lage Ca\~nellas$^{\star}$, Miguel Bordallo L\'opez$^{\star}$\end{tabular}}

\address{
$^{\star}$Center for Machine Vision and Signal Analysis (CMVS), University of Oulu, Finland \\
$^{\dagger}$Candour Ltd, Oulu, Finland \\
}

\begin{document}
%
\maketitle
\begin{abstract}
Video-based respiratory rate (RR) estimation is often unreliable due to inconsistent signal quality across extraction methods. We present a predictive, quality-aware framework that integrates heterogeneous signal sources with dynamic assessment of reliability. Ten signals are extracted from facial remote photoplethysmography (rPPG), upper-body motion, and deep learning pipelines, and analyzed using four spectral estimators: Welch’s method, Multiple Signal Classification (MUSIC), Fast Fourier Transform (FFT), and peak detection. Segment-level quality indices are then used to train machine learning models that predict accuracy or select the most reliable signal. This enables adaptive signal fusion and quality-based segment filtering. Experiments on three public datasets (OMuSense-23, COHFACE, MAHNOB-HCI) show that the proposed framework achieves lower RR estimation errors than individual methods in most cases, with performance gains depending on dataset characteristics. These findings highlight the potential of quality-driven predictive modeling to deliver scalable and generalizable video-based respiratory monitoring solutions. 
\end{abstract}



%
%

\section{Introduction}
\label{sec:intro}

Respiration is a fundamental vital sign, and monitoring RR is essential to assess patient health, detect respiratory distress, and prevent clinical deterioration \cite{nicolo2020importance}. Traditional RR monitoring relies on contact-based sensors such as impedance pneumography or capnography, which can be obtrusive, cause skin irritation, and limit long-term home monitoring or use with sensitive populations. Contactless monitoring using standard RGB cameras offers an accessible alternative, exploiting subtle respiration-induced phenomena such as skin color variations (rPPG) or thoraco-abdominal motion \cite{nguyen2025}. Such methods reduce infection risk, enable telemedicine, and have shown feasibility in hospitals and home environments. However, video-based respiration signals are extremely weak and susceptible to noise from motion, illumination changes, camera perspective, and segmentation errors, yielding unreliable RR estimates \cite{egorov2025gaze}.

Approaches to address these challenges follow two main directions. Algorithmic methods, such as CHROM \cite{haan2013chrom} and POS \cite{wang2017pos}, exploit color-space projections and temporal normalization to mitigate motion and illumination artifacts. Learning-based models, including ContrastPhys \cite{sun2022contrastphys} and PhysFormer \cite{zitong2022physformer}, infer physiological waveforms directly from video. Additional strategies include motion/color magnification and depth-based chest displacement \cite{manuel2024omusense}. Despite these advances, most studies optimize a single extractor and report average errors, with limited attention to per-segment reliability or principled fusion across multiple methods. Signal quality assessment is particularly underdeveloped for respiration: while contact PPG has extensive signal quality indices (SQIs) \cite{mohagheghian2022optimized}, camera-based metrics remain heterogeneous \cite{elgendi2024optimal}, and predictive quality control is rarely exploited.

To fill this gap, we propose a quality-aware framework for video-derived RR estimation. Our system concurrently runs ten extraction methods spanning rPPG, upper-body motion, and deep learning. For each signal segment, a set of SQIs is computed and fed to machine learning models to predict estimation error or identify the best-performing method. This enables dynamic fusion by selecting the optimal signal per window or filtering unreliable segments. Our contributions are: (i) incorporation of MUSIC with lagged autocorrelation matrix as a spectral estimator for  RR estimation; (ii) comparative analysis of ten respiratory extractors and four spectral estimators across three public datasets; (iii) a quality-aware framework for dynamic method fusion and low-quality segment filtering using ten SQIs. For context, we also report ground-truth (GT)-based reference ceilings that illustrate the achievable upper bound; these references are not available at inference time.

\section{Methodology}
\label{sec:method}

\begin{figure*}[ht!]
    \begin{center}
    \includegraphics[width=0.96\linewidth]{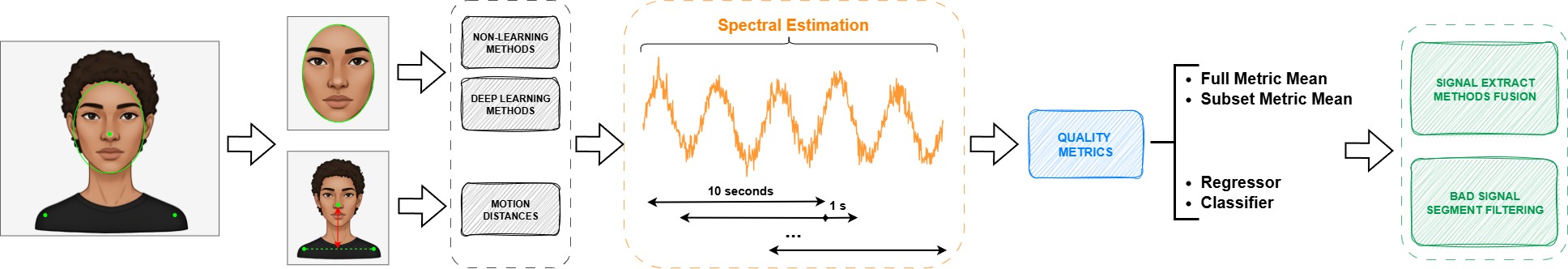}
    \end{center}
    \vspace{-5mm}
    \caption{Video-Based Respiratory Quality-aware Framework}
    \label{fig:pipeline}
    \vspace{-6mm}
\end{figure*}

We propose a framework for video-based respiratory signal estimation, illustrated in Fig.~\ref{fig:pipeline}, which concurrently applies ten complementary extraction methods, including rPPG-based non-learning and deep learning approaches, as well as upper-body motion, across three datasets. Each extracted signal is analyzed using four spectral estimators to derive RR estimates. For every signal segment, a comprehensive set of SQIs is computed and either aggregated into a composite score or used as input to machine learning models. These models predict either the segment-wise estimation MAE error or the index of the extraction method with the lowest MAE for that segment. Leveraging these predictions, the framework supports dynamic signal fusion, selecting the optimal signal at each time step, and quality-based filtering, discarding unreliable segments to improve the robustness of video-derived RR estimation.

\subsection{Signal Extraction}
Physiological signals were extracted from RGB videos using Mediapipe facial landmarks \cite{mediapipe} and upper-body keypoints. Facial landmark–based regions of interest (72×72) were used for rPPG-based RR estimation, while motion-based RR estimation relied on Euclidean distances between the nose and the shoulder midpoint. Three categories of methods were investigated: (i) four non-learning rPPG-based approaches: CHROM \cite{haan2013chrom}, GREEN \cite{verkruysse2008green}, OMIT \cite{casado2023omit}, and POS \cite{wang2017pos} which applied bandpass filtering to capture respiration-induced modulations; (ii) one motion-based approach based on inter-landmark distances \cite{mona2023distance}; and (iii) five deep learning rPPG-based approaches: iBVPNet \cite{joshi2024ibvp}, PhysNet \cite{zitong2019physnet}, PhysFormer \cite{zitong2022physformer}, and ContrastPhys \cite{sun2022contrastphys} are trained to directly extract respiratory signals from video data.

Deep learning models were trained using 5-fold cross-validation in the OMuSense-23 training set, with partial support from the rPPG-toolbox \cite{liu2023toolbox}. The best-performing validation model was applied to the OMuSense-23 \cite{manuel2024omusense} test set, as well as to unseen videos from COHFACE \cite{heusch2017cohface} and MAHNOB-HCI \cite{soleymani2012mahnob}, to evaluate cross-dataset generalization. GT signals were downsampled to match the video frame rate, and both GT and extracted signals were segmented using a 10-second sliding window with 1-second steps (9-second overlap). The signals were bandpass filtered between 0.1 and 0.5 Hz (6 to 30 breaths per minute) before respiratory rate estimation.

\subsection{Spectral Estimation}
\label{subsec:spectral}
Respiratory signals were analyzed using four spectral estimators: Welch, FFT, MUSIC, and Peak for RR estimation. Among these, MUSIC \cite{uysal2018music} was used as a high-resolution spectral estimator. In this study, a single respiratory signal was used to construct a lagged autocorrelation matrix, allowing it to operate in a single-channel setting rather than requiring multiple signals.We set the order of the model to \( p=2 \), which allocates two signal eigenvectors for a real sinusoid, and construct the Toeplitz autocorrelation matrix using \( M=2p=4\) lags ($k=0,1,2,3$). Eigenvalue decomposition separates the signal and noise subspaces, where the \( M-p \) smallest eigenvectors define the noise subspace. The steering vector across the frequency range is projected onto this subspace, and the MUSIC pseudo-spectrum is computed as the inverse of the projection, with peaks indicating dominant respiratory frequencies, as depicted in Algorithm~\ref{alg:music}. In downstream evaluations, \emph{Baseline} denotes the single best-performing signal and spectral estimator for each scenario: all methods (ALL), non-learning methods only (NLM), and deep learning methods only (DLM) on the given dataset.

\begin{algorithm}[ht!]
\caption{MUSIC Spectrum Estimation}
\label{alg:music}
\small 
\begin{algorithmic}[1]
\Require Signal $x[n]$, $n=0,\dots,N-1$; Sampling freq $f_s$; Model order $p$; FFT points $n_{\text{fft}}$
\Ensure Frequencies $f$; MUSIC pseudo-spectrum $P_{\text{MUSIC}}(f)$
\State $M \gets 2 \cdot p$
\State $R[k] \gets \frac{1}{N} \sum_{n=0}^{N-k-1} x[n] x[n+k],\ k=0,\dots,M-1$
\State $\mathbf{R} \gets \text{Toeplitz}(R)$
\State Eigen-decomposition: $\mathbf{R} \mathbf{v}_i = \lambda_i \mathbf{v}_i$
\State Noise subspace: $\mathbf{E}_{\text{noise}} \gets [\mathbf{v}_1, \dots, \mathbf{v}_{M-p}]$
\State $f_i \gets \frac{i-1}{n_{\text{fft}}} \frac{f_s}{2},\ i=1,\dots,n_{\text{fft}}$
\State $\mathbf{A}_{mi} \gets e^{-j 2 \pi m f_i / f_s},\ m=0,\dots,M-1$
\State $P_{\text{MUSIC}}(f_i) \gets \frac{1}{\sum_{k=0}^{M-p-1} |(\mathbf{E}_{\text{noise}}^H \mathbf{A})_{ki}|^2}$
\State \Return $f, P_{\text{MUSIC}}$
\end{algorithmic}
\end{algorithm}

\vspace{-3mm}
\subsection{Quality Metrics}
Signal quality was assessed using a set of metrics that capture both the quality and temporal characteristics of respiratory signals, including Zero Crossing Rate (ZCR), Hjorth parameters (mobility and complexity, as Hjorth-M and Hjorth-C, respectively), Signal-to-Noise Ratio (SNR), Irrelevant Power Ratio (IPR), Band Power Ratio (BPR), kurtosis (KURT), skewness (SKEW), Periodicity Index (PI), and Temporal Mean Cross-Correlation (TMCC) \cite{vallat2019quality,gideon2021ipr,nardelli2020bpr,elgendi2024optimal,charton2025}. Because metrics differ in scale and optimal direction, study-specific preferences were set: lower values for ZCR, Hjorth-M, and IPR; higher values for Hjorth-C, SNR, BPR, PI, and TMCC; and values near zero for SKEW and KURT. All metrics were then mapped so that lower values indicate higher quality and normalized to the 0–1 range for consistent aggregation.

We used normalized individual metrics to design two aggregate quality metrics: (i) The Full Metric Mean (FMM) was computed by averaging all normalized quality metrics; (ii) Subset Metric Mean (SMM): the mean of an optimal subset of metrics. All subsets of two or more metrics were evaluated, and for each subset, metrics were averaged at each time point with the extraction method yielding the lowest mean score selected. The subset minimizing overall MAE against GT was used to calculate SMM. In our experiments, we used Trainset-SMM, where the optimal subset was determined from quality metrics of ten extraction methods on the 5-fold OMuSense Trainset and then applied to other datasets and scenarios.


For predictive modeling, all quality metrics were first standardized via z-scores. We used lazypredict \cite{lazypredict} to evaluate multiple regressors and classifiers, from which the two best of each type were selected. Regressors (iii: ExtraTrees and Multi-Layer Perceptron, MLP) directly predicted MAE errors, while classifiers (iv: LightGBM and the XGBoost Classifier, XGBC) predicted the index of the extraction method with the lowest MAE. The four strategies (i-iv) were applied for intra- and cross-dataset signal fusion and compared to the \emph{Baseline} across three evaluation scenarios. Additionally, FMM with the MLP regressor was used to filter low-quality segments.


We report two GT-based reference lines as  \emph{oracle upper bounds}: (a)  \emph{GT-SMM}, where the optimal metric subset is selected using GT to minimize MAE per dataset/scenario; (b)  \emph{GT MAE}, where GT MAE errors are used to select the best method or discard low-quality segments.

\section{Results}
\label{results}


\begin{figure}[ht!]
  \centering
  \includegraphics[width=\linewidth]{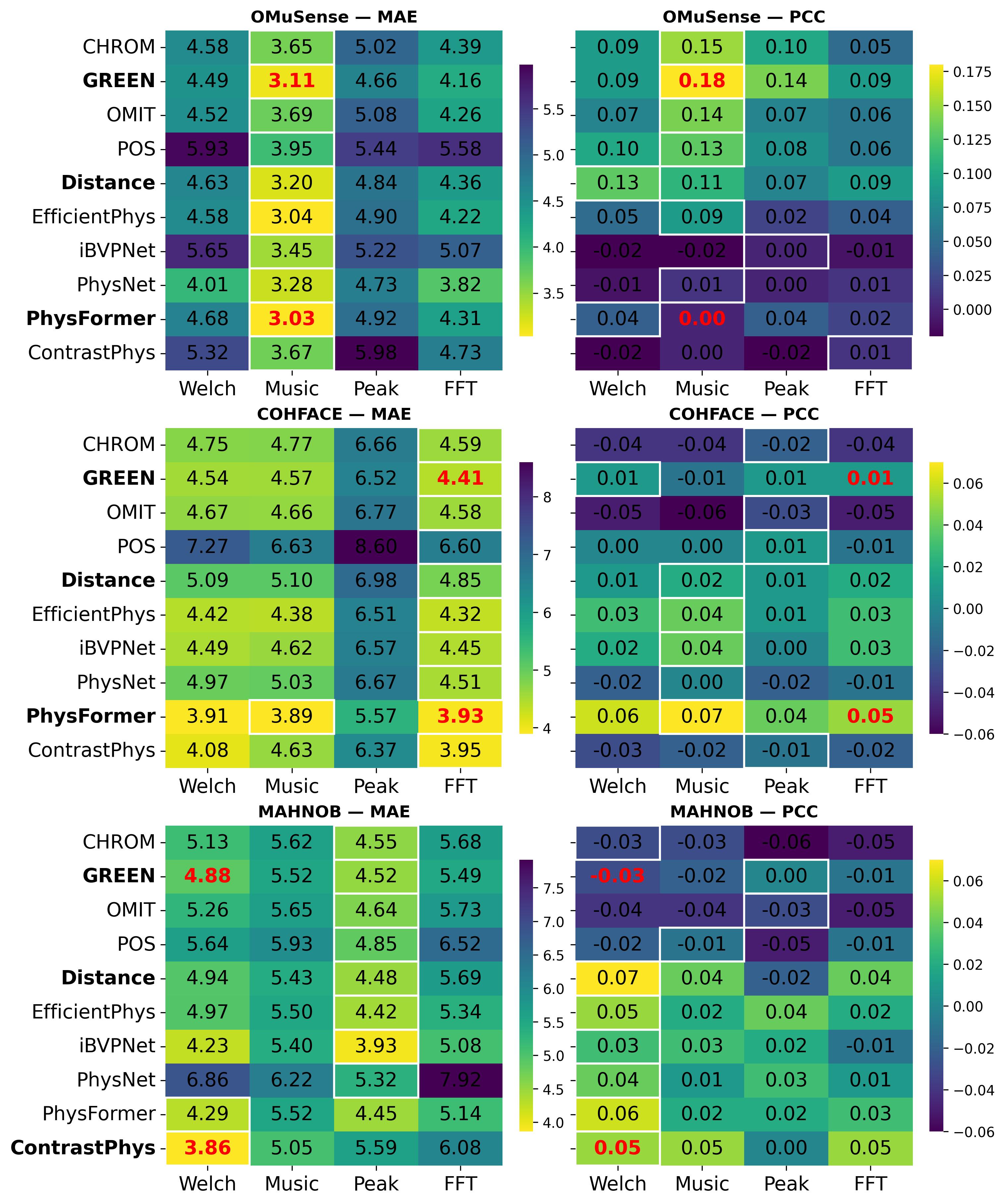}
  \vspace{-5mm}
  \caption{RR Estimation Heatmaps: columns = spectral estimators, rows = extraction methods; lower MAE / higher PCC is better, white boxes = best, bold red = baseline.}
  \label{fig:spectral_comp_matrix}
\end{figure}



We compare four spectral estimators across signals from ten extractors to assess how the spectral stage affects RR accuracy, aiming to identify a stable baseline for subsequent fusion and filtering and to reveal dataset-dependent effects that may bias downstream analyses. As shown in Fig.~\ref{fig:spectral_comp_matrix}, MUSIC performs best on OMuSense, FFT on COHFACE, and Peak shows the lowest errors on MAHNOB but is unsuitable for frequency-based quality metrics, so Welch is used as the baseline for subsequent evaluations. In the figure, red-bold values denote the \emph{Baseline} for signal fusion across three scenarios, and three signal extraction methods: one non-learning (GREEN), one motion-based (Distance), and one deep learning (PhysFormer or ContrastPhys) are highlighted for quality-based filtering, each using the aforementioned spectral estimators.

Table~\ref{tab:qualitymetric} shows that the optimal subset of quality metrics varies across datasets and scenarios. Nonetheless, Hjorth-M, Hjorth-C, IPR, and BPR are the most consistently selected and robust, while ZCR, SNR, and PI provide complementary information depending on the dataset and method. NLM often favor Hjorth-M, BPR, and PI, whereas DLM consistently include Hjorth-M and BPR. 


\begin{table}[h!]
\centering
\def\arraystretch{1.0}
\setlength{\tabcolsep}{1.5em} 
\resizebox{\linewidth}{!}{
\begin{tabular}{lcl}
\hline
\textbf{Dataset} & \textbf{Scenario} & \textbf{Subset Metric}\\
\hline
\multirow{3}{*}{\begin{tabular}[l]{@{}l@{}}OmuSense \\Trainset\end{tabular}}
& ALL   & Hjorth-M, Hjorth-C, IPR, BPR, PI \\
& NLM   & Hjorth-M, BPR, PI \\
& DLM   & Hjorth-M, BPR, PI \\
\hline
\multirow{3}{*}{\begin{tabular}[l]{@{}l@{}}OmuSense \\Testset\end{tabular}}
  & ALL  & Hjorth-M, Hjorth-C, IPR, BPR        \\
  & NLM  & ZCR, IPR, BPR                    \\
  & DLM  & Hjorth-M, BPR                  \\
\hline
\multirow{3}{*}{COHFACE} 
  & ALL  & ZCR, Hjorth-M, Hjorth-C, SNR, IPR    \\
  & NLM  & ZCR, Hjorth-M, Hjorth-C              \\
  & DLM  & Hjorth-M, Hjorth-C, IPR, PI          \\
\hline
\multirow{3}{*}{MAHNOB} 
  & ALL  & Hjorth-C, SNR, BPR                \\
  & NLM  & SNR, BPR                         \\
  & DLM  & Hjorth-M, SNR, IPR, BPR, TMCC              \\
\hline
\end{tabular}
}
\caption{Optimal Subset of Metrics (Lowest GT MAE).}
\label{tab:qualitymetric}
\vspace{-3mm}
\end{table}

\begin{table*}[!t]
\centering
\def\arraystretch{1.05}
\setlength{\tabcolsep}{0.6em} 
\resizebox{\linewidth}{!}{
\begin{tabular}{l *{6}{c}|*{6}{c}|*{6}{c}}
\hline

\multicolumn{1}{c}{} & \multicolumn{6}{c|}{OmuSense Testset} & \multicolumn{6}{c|}{COHFACE} & \multicolumn{6}{c}{MAHNOB} \\
\cline{2-7} \cline{8-13}  \cline{14-19}

\multirow{2}{*}{\begin{tabular}[c]{@{}c@{}}Fusion\\Meth.\end{tabular}}
& \multicolumn{2}{c}{ALL}
& \multicolumn{2}{c}{NLM} 
& \multicolumn{2}{c|}{DLM}

& \multicolumn{2}{c}{ALL}
& \multicolumn{2}{c}{NLM} 
& \multicolumn{2}{c|}{DLM}

& \multicolumn{2}{c}{ALL}
& \multicolumn{2}{c}{NLM} 
& \multicolumn{2}{c}{DLM}
\\
\cline{2-3} \cline{4-5} \cline{6-7} \cline{8-9} \cline{10-11} \cline{12-13} \cline{14-15} \cline{16-17} \cline{18-19}

\multicolumn{1}{c}{}
& \multicolumn{1}{c}{MAE} & \multicolumn{1}{c}{PCC} 
& \multicolumn{1}{c}{MAE} & \multicolumn{1}{c}{PCC}  
& \multicolumn{1}{c}{MAE} & \multicolumn{1}{c|}{PCC}  

& \multicolumn{1}{c}{MAE} & \multicolumn{1}{c}{PCC} 
& \multicolumn{1}{c}{MAE} & \multicolumn{1}{c}{PCC}  
& \multicolumn{1}{c}{MAE} & \multicolumn{1}{c|}{PCC}  

& \multicolumn{1}{c}{MAE} & \multicolumn{1}{c}{PCC} 
& \multicolumn{1}{c}{MAE} & \multicolumn{1}{c}{PCC}  
& \multicolumn{1}{c}{MAE} & \multicolumn{1}{c}{PCC}  
\\
\hline

GT MAE & 0.44 & 0.93 & 1.00 & 0.82 & 0.99 & 0.80 
& 1.20 & 0.91 & 1.92 & 0.74 & 1.48 & 0.84 
& 0.75 & 0.94 & 1.80 & 0.73 & 1.29 & 0.86 \\

\B Baseline & \B 3.03 & \B 0.00 & \B 3.11 & \B 0.18 & \B 3.03 & \B 0.00 
& \B 3.93 & \B 0.05 & \B 4.41 & \B 0.01 & \B 3.93 & \B 0.00 
& \B 3.86 & \B 0.05 & \B 4.88 & \B -0.03 & \B 3.86 & \B 0.05 \\

FMM & 2.77 & 0.09 & 2.88 & 0.14 & 2.94 & 0.01 
& 4.28 & -0.02 & 4.63 & -0.02 & 4.01 & 0.01 
& 3.66 & 0.03 & 4.16 & 0.03 & 3.65 & 0.04 \\ 
\hline

GT-SMM & 2.62 & 0.10 & 2.67 & 0.16 & 2.56 & 0.04 
& 3.55 & 0.01 & 3.72 & -0.01 & 3.52 & 0.02 
& 3.34 & 0.06 & 3.64 & 0.02 & 3.48 & 0.08 \\

Trainset-SMM  & 2.63 & 0.09 & 2.68 & 0.15 & 2.57 & 0.03 
& 3.58 & -0.01 & 3.93 & -0.02 & 3.58 & 0.01 
& 4.16 & 0.03 & 4.22 & 0.02 & 3.85 & 0.00 \\

\hline
ExtraTrees & 3.18 & 0.08 & 3.11 & 0.14 & 3.01 & 0.02 
& 4.63 & 0.00 & 4.82 & -0.02 & 4.15 & 0.01 
& 4.80 & 0.01 & 4.75 & -0.02 & 4.68 & 0.05 \\

MLP & 2.37 & 0.14 & 2.40 & 0.15 & 2.51 & 0.03 
& 3.91 & 0.01 & 4.23 & -0.01 & 3.78 & 0.02 
& 4.00 & 0.00 & 4.16 & -0.01 & 4.27 & -0.09 \\

\hline
LGBM & 2.64 & 0.10 & 2.59 & 0.10 & 2.59 & 0.03 
& 4.14 & 0.00 & 4.27 & -0.02 & 3.80 & 0.02 
& 4.22 & 0.01 & 4.17 & -0.01 & 4.45 & -0.01 \\

XGBC & 2.67 & 0.09 & 2.60 & 0.13 & 2.63 & 0.04 
& 4.40 & -0.01 & 4.36 & 0.01 & 3.93 & 0.01 
& 4.26 & 0.00 & 4.20 & -0.01 & 4.53 & 0.01 \\
\hline
\addlinespace
\end{tabular}
}
\vspace{-3mm}
\caption{Respiratory Signal Fusion Methods. \textit{Note:} \emph{GT MAE} and \emph{GT-SMM} are oracle upper bounds.}

\label{tab:fusion}
\vspace{-4mm}
\end{table*}

Tables~\ref{tab:fusion} summarize fusion performance for RR estimation across datasets. Most methods reduce MAE relative to the baseline, showing the benefit of combining multiple signals or metrics, although improvements are dataset- and method-dependent. In some COHFACE and MAHNOB scenarios, ExtraTrees, LGBM, or XGBC fail to improve MAE. The Trainset-SMM derived from OMuSense training data performs close to the Testset and COHFACE GT-SMM, highlighting the potential of quality-aware fusion trained on representative data. MLP Regressor generally achieves the lowest MAE, but PCC gains remain limited, showing that temporal alignment is not always improved. Fusion reduces MAE in most cases; however, some scenarios show little to no improvement, and the results remain far from the GT MAE, underscoring the potential for further enhancement and the need for adaptive, dataset-aware strategies.

A novel insight is that across datasets, the dominant performance lever was not a new extractor or spectral method but learning when to trust which one: a compact, transferable SQI core (Hjorth-M, BPR, often IPR/PI) enables per-window selection that consistently outperforms the best single pipeline despite front-end and dataset shifts.





Table~\ref{tab:segment} shows the effect of low-quality segment filtering on RR estimation for three representative methods per dataset, applied incrementally from 0–50\% of predicted low-quality segments. GT selection achieves the largest MAE reduction, representing the upper bound as an \emph{oracle}, not deployable at test time. Both FMM and MLP-based filtering reduce MAE relative to unfiltered signals, with MLP generally outperforming FMM, particularly in OMuSense and MAHNOB. Improvements are dataset- and method-dependent: in COHFACE, FMM offers limited improvement for GREEN and Distance, while MLP provides more consistent reductions. PhysFormer shows a smaller increase, reflecting its robustness. Overall, predictive quality-based filtering can improve RR estimation, but effectiveness depends on the method and dataset, and filtering must balance error reduction with sufficient signal coverage.

\begin{table}[!h]
\centering
\setlength{\tabcolsep}{1.5pt} 
\renewcommand{\arraystretch}{0.8} 
\resizebox{\linewidth}{!}{
\begin{tabular}{l *{13}{c}}
\toprule 

\multirow{2}{*}{\begin{tabular}[c]{@{}c@{}}Filter\\Meth.\end{tabular}} & \multirow{2}{*}{\begin{tabular}[c]{@{}c@{}}rPPG \\Meth.\end{tabular}}
& \multicolumn{2}{c}{0 \%}
& \multicolumn{2}{c}{10 \%} 
& \multicolumn{2}{c}{20 \%}
& \multicolumn{2}{c}{30 \%}
& \multicolumn{2}{c}{40 \%}
& \multicolumn{2}{c}{50 \%} \\
\cmidrule(lr){3-4} \cmidrule(lr){5-6} \cmidrule(lr){7-8} \cmidrule(lr){9-10} \cmidrule(lr){11-12} \cmidrule(lr){13-14}

\multicolumn{1}{c}{}
& \multicolumn{1}{c}{}
& \multicolumn{1}{c}{MAE} & \multicolumn{1}{c}{PCC} 
& \multicolumn{1}{c}{MAE} & \multicolumn{1}{c}{PCC}  
& \multicolumn{1}{c}{MAE} & \multicolumn{1}{c}{PCC}  
& \multicolumn{1}{c}{MAE} & \multicolumn{1}{c}{PCC} 
& \multicolumn{1}{c}{MAE} & \multicolumn{1}{c}{PCC}  
& \multicolumn{1}{c}{MAE} & \multicolumn{1}{c}{PCC}  
\\
\midrule

\multicolumn{14}{c}{OmuSense Test} \\
\midrule

GT MAE & \multirow{3}{*}{GREEN} & \multirow{3}{*}{\B 3.11} & \multirow{3}{*}{\B 0.18} & 2.49 & 0.42 & 2.07 & 0.55 & 1.69 & 0.65 & 1.39 & 0.73 & 1.05 & 0.83 \\
FMM & &  &  & 2.96 & 0.16 & 2.88 & 0.15 & 2.82 & 0.15 & 2.76 & 0.15 & 2.70 & 0.15 \\
MLP  & &  &  & 2.79 & 0.16 & 2.61 & 0.15 & 2.48 & 0.15 & 2.37 & 0.15 & 2.29 & 0.15 \\
\midrule

GT MAE & \multirow{3}{*}{Distance} & \multirow{3}{*}{\B 3.20} & \multirow{3}{*}{\B 0.11} & 2.56 & 0.35 & 2.12 & 0.50 & 1.74 & 0.61 & 1.44 & 0.71 & 1.07 & 0.82 \\
FMM & &  &  & 3.04 & 0.12 & 2.99 & 0.12 & 2.96 & 0.12 & 2.92 & 0.12 & 2.87 & 0.12 \\
MLP  & &  &  & 2.84 & 0.13 & 2.64 & 0.14 & 2.51 & 0.13 & 2.41 & 0.14 & 2.33 & 0.13 \\
\midrule

GT MAE & \multirow{3}{*}{\begin{tabular}[c]{@{}c@{}}Phys\\Former\end{tabular}}
& \multirow{3}{*}{\B 3.03} & \multirow{3}{*}{\B 0.00} & 2.41 & 0.65 & 2.00 & 0.42 & 1.65 & 0.55 & 1.34 & 0.67 & 1.02 & 0.78 \\
FMM & &  &  & 2.96 & 0.00 & 2.91 & 0.01 & 2.88 & 0.00 & 2.83 & 0.01 & 2.79 & 0.01 \\
MLP  & &  &  & 2.76 & 0.05 & 2.61 & 0.07 & 2.50 & 0.08 & 2.41 & 0.10 & 2.30 & 0.12 \\

\midrule

\multicolumn{14}{c}{COHFACE} \\
\midrule

GT MAE & \multirow{3}{*}{GREEN} & \multirow{3}{*}{\B 4.41} & \multirow{3}{*}{\B 0.01} & 3.54 & 0.19 & 2.86 & 0.41 & 2.30 & 0.57 & 1.86 & 0.68 & 1.45 & 0.80 \\
FMM & &  &  & 4.40 & 0.01 & 4.41 & 0.00 & 4.38 & 0.00 & 3.33 & -0.01 & 4.26 & 0.00 \\
MLP  & &  &  & 4.22 & 0.01 & 4.11 & 0.00 & 4.00 & 0.00 & 3.96 & 0.00 & 3.90 & 0.01 \\
\midrule

GT MAE & \multirow{3}{*}{Distance} & \multirow{3}{*}{\B 4.85} & \multirow{3}{*}{\B 0.02} & 3.88 & 0.23 & 3.19 & 0.43 & 2.59 & 0.60 &2.16 & 0.69 & 1.75 & 0.78 \\
FMM & &  &  & 4.81 & 0.02 & 4.72 & 0.03 & 4.64 & 0.03 & 4.62 & 0.03 & 4.52 & 0.02 \\
MLP  & &  &  & 4.56 & 0.03 & 4.38 & 0.03 & 4.30 & 0.03 & 4.23 & 0.03 & 4.16 & 0.03 \\
\midrule

GT MAE & \multirow{3}{*}{\begin{tabular}[c]{@{}c@{}}Phys\\Former\end{tabular}}
& \multirow{3}{*}{\B 3.93} & \multirow{3}{*}{\B 0.05} & 3.15 & 0.28 & 2.54 & 0.48 & 2.14 & 0.59 & 1.75 & 0.70 & 1.37 & 0.82 \\
FMM & &  &  & 3.90 & 0.04 & 3.88 & 0.03 & 3.88 & 0.02 & 3.90 & 0.02 & 3.91 & 0.00 \\
MLP  & &  &  & 3.77 & 0.03 & 3.69 & 0.03 & 3.64 & 0.02 & 3.60 & 0.02 & 3.59 & 0.01 \\

\midrule

\multicolumn{14}{c}{MAHNOB} \\
\midrule

GT MAE & \multirow{3}{*}{GREEN} & \multirow{3}{*}{\B 4.88} & \multirow{3}{*}{\B -0.03} & 4.05 & 0.19 & 3.41 & 0.34 & 2.84 & 0.49 & 2.35 & 0.62 & 1.88 & 0.73 \\
FMM & &  &  & 4.68 & -0.03 & 4.50 & -0.03 & 4.39 & -0.03 & 4.26 & -0.04 & 4.05 & -0.06 \\
MLP & &  &  & 4.67 & -0.03 & 4.52 & -0.03 & 4.39 & -0.04 & 4.32 & -0.05 & 4.18 & -0.07 \\
\midrule

GT MAE & \multirow{3}{*}{Distance} & \multirow{3}{*}{\B 4.94} & \multirow{3}{*}{\B 0.07} & 4.11 & 0.28 & 3.49 & 0.42 & 2.93 & 0.55 & 2.45 & 0.65 & 2.01 & 0.74 \\
FMM & &  &  & 4.69 & 0.07 & 4.49 & 0.07 & 4.33 & 0.05 & 4.22 & 0.05 & 3.97 & 0.06 \\
MLP & &  &  & 4.75 & 0.05 & 4.62 & 0.04 & 4.46 & 0.05 & 4.45 & 0.04 & 4.30 & 0.05 \\
\midrule

GT MAE & \multirow{3}{*}{\begin{tabular}[c]{@{}c@{}}Contrast\\Phys\end{tabular}}
& \multirow{3}{*}{\B 3.86} & \multirow{3}{*}{\B 0.05} & 3.12 & 0.27 & 2.62 & 0.42 & 2.19 & 0.55 & 1.83 & 0.67 & 1.48 & 0.76 \\
FMM & &  &  & 3.54 & 0.05 & 3.50 & 0.04 & 3.47 & 0.03 & 3.50 & 0.02 & 3.53 & 0.02 \\
MLP & &  &  & 3.66 & 0.04 & 3.52 & 0.05 & 3.50 & 0.05 & 3.49 & 0.05 & 3.53 & 0.03 \\
\bottomrule
\addlinespace
\end{tabular}
}
\caption{Impact of Low-Quality Segment Filtering (\%). \textit{Note:} Rows \emph{GT MAE} are oracle upper bounds.}
\label{tab:segment}
\vspace{-4mm}
\end{table}

\vspace{-3mm}
\section{Conclusion}
\label{concl}

This study presents a quality-aware framework for respiratory rate estimation from RGB videos, integrating non-learning, deep learning rPPG, and motion-based methods with diverse spectral estimators and machine learning–based quality assessment. Dynamic signal fusion and low-quality segment filtering consistently reduced MAE across datasets. The subset metric analysis identified a small set of indices that effectively guide quality-aware processing, although the optimal subset varies with the dataset and scenario. While MAE improved substantially, temporal correlation with PCC remained limited, reflecting the challenge of capturing fine-grained respiratory dynamics. These results demonstrate the potential of predictive quality assessment and adaptive fusion, with future work targeting temporal fidelity, cross-dataset generalization, and real-time applications. Practically, our results show that a lightweight learned controller yields larger gains than adding new extractors, shifting the focus to temporal consistency and dataset-aware spectral tuning.

\textbf{Acknowledgments} The research was supported by the Business Finland WISEC project (Grant 3630/31/2024), the University of Oulu and the Research Council of Finland (former Academy of Finland) 6G Flagship Programme (Grant Number: $346208$), Profi5 HiDyn programme (326291), and Profi7 Hybrid intelligence program (352788).

\bibliographystyle{IEEEbib}
\bibliography{strings,references}

\end{document}